\documentstyle{article}
\def\report{536-96}
\title{Nonuniform Markov Models%
\thanks{Thanks to Andrew Appel, Joe Kupin and Harry Printz for their
critique.  Our implementation of the nonuniform model used the library
of practical abstractions (Ristad and Yianilos, 1996).  Both authors
are partially supported by Young Investigator Award IRI-0258517 to the
first author from the National Science Foundation.}}
\author{Eric Sven Ristad \and Robert G. Thomas}
\date{November 1996}

\def\shat{\hat{s}}
\def\pbar{\bar{p}}
\def\jmax{j_{\rm max}}
\def\magnitude#1{\left| #1 \right|}
\def\tuple<#1>{\langle #1 \rangle}
\newtheorem{theorem}{Theorem}
\newcommand{\proof}{\noindent{\bf Proof. }}
\newcommand{\figurerule}{\par\vskip 5pt\hrule\vskip 5pt\relax}
\def\qed{\large ~$\Box$ \normalsize}

\begin{document}

\makeatletter
\def\abstract{ \vfil
\begin{center}{\bf Abstract}\end{center}}
\begin{titlepage}
\let\footnotesize\small \let\footnoterule\relax
\leavevmode
\vskip .7in
\begin{center}
{\Large\bf \@title \par}
\vskip 1pc
\begin{tabular}[t]{c}\@author 
\end{tabular}
 
\vskip 1pc
Research Report CS-TR-\report\par
\@date

\end{center}
\noindent
\@thanks

\begin{abstract}
A statistical language model assigns probability to strings of
arbitrary length.  Unfortunately, it is not possible to gather
reliable statistics on strings of arbitrary length from a finite
corpus.  Therefore, a statistical language model must decide that each
symbol in a string depends on at most a small, finite number of other
symbols in the string.  In this report we propose a new way to model
conditional independence in Markov models.  The central feature of our
nonuniform Markov model is that it makes predictions of varying
lengths using contexts of varying lengths.  Experiments on the Wall
Street Journal reveal that the nonuniform model performs slightly
better than the classic interpolated Markov model.  This result is
somewhat remarkable because both models contain identical numbers of
parameters whose values are estimated in a similar manner.  The
only difference between the two models is how they combine the
statistics of longer and shorter strings.
\end{abstract}
\vskip 1pc

\noindent {\bf Keywords:}
nonuniform Markov model,
interpolated Markov model,
conditional independence,
statistical language model,
discrete time series.
\end{titlepage}

\section{Introduction}

The task of statistical language modeling is to accurately predict the
future utterances of a language user.  The probability that a given
language user will produce a given utterance at a given moment depends
on the language user's knowledge of language and of the world.  Our
current understanding of the language user's cognitive abilities is
too impoverished for us to build plausible models of the language
user's knowledge, and so we must be content to model the observables
as best we can.  Here the observables are the word sequences produced
by language users.  And so our goal is to assign accurate
probabilities to word sequences.

The interpolated Markov model \cite{jelinek-mercer:80} and its cousin
the backoff model \cite{cleary-witten:84,katz:87,ristad-thomas:acl95}
have long been the workhorses of the statistical language modeling
community.  These traditional models rely only on the frequencies of
strings up to a fixed length.  Recent research in statistical language
modeling has focused primarily on developing more powerful model
classes \cite{jelinek-etal:92a,lafferty-etal:92} as well as on adding
new sources of information to the traditional models
\cite{lau-etal:93,della-pietra-etal:94}.  In contrast, the goal of
this work is to find a more effective way to use the statistics of
finite length strings.  The distinguishing feature of our model is
that it acquires beliefs about conditional independence, and uses
those beliefs to make predictions of varying lengths using contexts of
varying lengths.

We believe that our work has two contributions to offer to the field
of Markov modeling.  The first contribution is our interpretation of
the interpolation parameters as beliefs about conditional
independence.  Prior work on interpolated Markov models has
interpreted the interpolation parameters as smoothing the ``specific
probabilities'' with the ``general probabilities''
\cite{jelinek-mercer:80,mackay-peto:94}.  Our interpretation gives
rise to the second contribution of our work, namely, a class of {\it
nonuniform\/} Markov models that make predictions of varying lengths
using contexts of varying lengths.  Nonuniform predictions is a
principled way to perform alphabet extension, that is, to make a
string become a symbol in the alphabet, an {\it ad hoc\/} technique
that can improve model performance \cite{jeanrenaud-etal:95}.

The remainder of this report is organized into four sections.  In
section~\ref{model-section} we motivate the nonuniform model as
arising from the proper generative interpretation of our beliefs about
conditional independence.  In section~\ref{algorithms-section} we
provide efficient algorithms for evaluating the probability of a
string according to a nonuniform model, for finding the most likely
nonuniform generation path for a given string, and for optimizing the
parameters of a nonuniform model on a training corpus.  Finally, in
section~\ref{empirical-section} compare the performance of the classic
interpolated Markov model and the nonuniform model on the Wall Street
Journal.  The nonuniform model performs slightly better than the
classic model under equivalent experimental conditions.  This result
is somewhat remarkable, since the only difference between these two
models is how they interpret the interpolation parameters.

\section{Nonuniform Model}\label{model-section}

A statistical language model assigns probability to strings of
arbitrary length.  Unfortunately, it is not possible to gather
reliable statistics on strings of arbitrary length from a finite
corpus.  In practice, this difficulty is quite severe.  There are
$k^n$ logically possible strings of length $n$ over an alphabet of
size $k$, but there are at most $T-n+1$ distinct strings of length $n$
in a corpus of length $T$.  Nearly all of the $n$-grams do not occur
in any finite corpus, and of the $n$-grams that do occur, nearly all
occur only once.  Therefore, we must decide that each symbol in a
string depends only on at most a small, finite number of other symbols
and is conditionally independent of all other symbols in the string.

For example, a Markov model of order $n$ stipulates that each symbol
depends only on the $n$ most recent symbols, and is conditionally
independent of all other past symbols,
\begin{displaymath}
	p(x_i|x_1\ldots x_{i-1}) \doteq p(x_i|x_{i-n}\ldots x_{i-1})
\end{displaymath}
where the probability $p(x^T|T)$ of a string $x^T$ of length $T$ is
then calculated as a product of $T$ conditional probabilities.
\begin{displaymath}
\begin{array}{rcl}
	p(x^T|T) & = & \prod_{i=1}^{T} p(x_i | x_1 \ldots x_{i-1}, T) \\
	\displaystyle
		 & \doteq & \prod_{i=1}^{T} p(x_i | x_{i-n}\ldots x_{i-1}, T)
\end{array}
\end{displaymath}

We are trying to model the observable correlates of a cognitive
process far more complex and powerful than a fixed order Markov model.
Consequently, we cannot afford to take such a simple-minded approach
to conditional independence.  Rather than stipulate the point of
conditional independence {\it a priori\/}, as in a Markov model, we
would like our model to acquire beliefs about conditional independence
based on empirical evidence.

In this section, we provide three different generative interpretations
for the state-conditional interpolation parameters of a Markov model.
These interpretations give rise to an interpolated context model, an
interpolated state model, and our nonuniform model.  Next, we compare
the ability of these three interpretations to model local independence
and global independence.  We argue that the nonuniform model combines
the ability of the state model to properly model global independence
with the ability of the context model to properly model local
independence.  Finally, we prove that the nonuniform model is
fundamentally different from the other two models because it is not
possible to map a nonuniform model into an extensionally equivalent
context or state model.

Let us first define our notation.  Let $A$ be a finite alphabet of
distinct symbols, $\magnitude{A} = k$, and let $x^T \in A^T$ denote an
arbitrary string of length $T$ over the alphabet $A$.  Then $x_i^j$
denotes the substring of $x^T$ that begins at position $i$ and ends at
position $j$.  For convenience, we abbreviate the unit length
substring $x_i^i$ as $x_i$ and the length $t$ prefix of $x^T$ as
$x^t$.

 \subsection{Three Interpolated Models}

An interpolated Markov model $\phi = \tuple<n,A,\delta,\lambda>$
consists of a maximal string length $n$, a finite alphabet $A$, a set
of string probabilities $\delta: A^{\leq n} \rightarrow [0,1]$, and
the interpolation parameters $\lambda: A^{<n}\rightarrow [0,1]$.
Given a string $y^l$, $l<n$, the string probabilities $\delta(y^l)$
are typically their empirical probabilities in a training corpus.  The
only difference between our three models will be how the interpolation
parameters $\lambda$ are interpreted.

Let us now consider three generative interpretations of the
interpolated Markov model: the context model, the state model, and our
nonuniform model.  A context model interprets the $\lambda$ parameters
as combining the predictions from Markov models of varying orders.  A
state model interprets the $\lambda$ parameters as hidden transitions
from a higher order Markov model to a lower order Markov model.  The
state and context models are both uniform models because they always
predict unit-length strings.  A nonuniform model interprets the
$\lambda$ parameters as beliefs about conditional independence.

In each case, we let $\pbar_c(i|x_{t-m+1}^t)$ be the probability that
we pick a context of length $i$ in the history $x_{t-m+1}^t$ and let
$\pbar_v(y_1^j|x_{t-i+1}^t)$ be the probability that we make a
prediction $y_1^j$ of length $j$ in the chosen context $x_{t-i+1}^t$.

  \subsubsection{Context Model}

In the interpolated context model, the interpolation parameters are
understood as smoothing the conditional probabilities estimated from
longer histories with those estimated from shorter histories
\cite{jelinek-mercer:80,mackay-peto:94}.  Longer histories support
stronger predictions, while shorter histories have more accurate
statistics.  Interpolating the predictions from histories of different
lengths results in more accurate predictions than can be obtained from
any fixed history length.  This interpretation of the interpolation
parameters was originally proposed by Jelinek and Mercer
\cite{jelinek-mercer:80}.  It leads to the following generation
algorithm, where the hidden transition from a longer context to a
shorter context (line 3) is temporary, used only for the current
prediction (line 4).

{\sf
\begin{tabbing}
aaa \= aaa \= aaa \= a \= \kill
{\sc context-generate}($T$,$\phi$) \\
1. \> Initialize $t := 0$; $x_1^0 := \epsilon$; \\
2. \> Until $t = T$ \\
3. \> \> Pick context length $i$ in [$0,\min(t,n-1)$] \\
   \> \> \> $\pbar_c(i|x^t)$ = 
		$\lambda(x_{t-i+1}^t)
			\prod_{l=\min(t,n-1)}^{i+1} (1 - \lambda(x_{t-l+1}^t))$ \\
4. \> \> Make one symbol prediction $y^1$ \\
   \> \> \> $\pbar_v(y^1|x_{t-i+1}^t) = \delta(y^1|x_{t-i+1}^t, i+1)$ \\
5. \> \> Extend history $x_1^t$ by prediction $y^1$\\
   \> \> \> $x_1^{t+1} := x_1^t y^1$; $t := t + 1$; \\
6. \> return($x^T$); \\
\end{tabbing}
}

The probability $p_c(x_i|x^{i-1},\phi)$ assigned by an interpolated
context model $\phi$ to a symbol $x_i$ in the history $x^{i-1}$ has a
particularly simple form (\ref{context-evaluate}),
\begin{equation}\label{context-evaluate}
p_c(x_i|x^{i-1},\phi) = 
	\lambda(x^{i-1})\delta(x_i|x^{i-1})
	+ (1-\lambda(x^{i-1})) p_c(x_i|x_2^{i-1},\phi)
\end{equation}
where $\lambda(x^i) = 0$ for $i \geq n$ and $\lambda(\epsilon) = 1$.

  \subsubsection{State Model}

Alternately, the interpolation parameters may be understood as
modeling our beliefs about how much of the past is necessary to
predict a state transition in an underlying Markov source of unknown
order.  This interpretation leads to the following generation
algorithm, where the hidden transition from a state of a higher order
model to a state of a lower order model (line 3) is permanent (line
4).

{\sf
\begin{tabbing}
aaa \= aaa \= aaa \= a \= \kill
{\sc state-generate}($T$,$\phi$) \\
1. \> Initialize $t := 0$; $x_1^0 := \epsilon$; $m := 0$; \\
2. \> Until $t = T$ \\
3. \> \> Pick context length $i$ in [$0,m$] \\
   \> \> \> $\pbar_c(i|x_{t-m+1}^t)$ = 
		$\lambda(x_{t-i+1}^t)$
			$\prod_{l=m}^{i+1} (1 - \lambda(x_{t-l+1}^t))$ \\
4. \> \> $m := i$; \\
5. \> \> Make one symbol prediction $y^1$ \\
   \> \> \> $\pbar_v(y^1|x_{t-i+1}^t) = \delta(y^1|x_{t-i+1}^t, i+1)$ \\
6. \> \> Extend history $x_1^t$ by prediction $y^1$\\
   \> \> \> $x^{t+1} := x^t y^1$; $t := t + 1$; $m := \min(m+1,n-1)$; \\
7. \> return($x^T$); \\
\end{tabbing}
}

  \subsubsection{Nonuniform Model}\label{model-subsection}

We develop the following model of conditional independence.  Let
$\iota(x^n)$ be our degree of belief that $x_n$ depends on $x_1$ in a
string $x_1^n$ of length $n$
\begin{displaymath}
  \iota(x^n) \doteq p( p(x_n|x_1\ldots x_{n-1}) \neq p(x_n|x_2\ldots x_{n-1}) )
\end{displaymath}
and let $\lambda(x^i)$ be our degree of belief that the next $n-i$
symbols depend on $x_1$, a kind of expected dependence.
\begin{displaymath}
  \lambda(x^i) \doteq \sum_{y^{n-i}} p(y^{n-i}| x^i) \iota(x^i y^{n-i})
\end{displaymath}
Our beliefs about independence are determined in large part by the
robustness of our statistics.  If we do not believe that our model
$\delta(\cdot|x^i)$ of the source state transition probabilities is
accurate, then our $\lambda(x^i)$ will be low.

Our beliefs about conditional independence have two implications.  The
first implication, as in the uniform model, is that we should
transition from a longer context $x^i$ to the shorter context $x_2^i$
with probability $1 - \lambda(x^i)$.  This expresses our belief of
degree $1 - \lambda(x^i)$ that the future does not depend on $x_1$.
The second implication, which is unique to the nonuniform model, is
that we should transition from a shorter prediction $y^{j-1}$ to a
longer prediction $y^j$ in the chosen context $x^i$ with probability
$\lambda(x^i y^{j-1})$.  This implication follows from our belief of
degree $\lambda(x^i y^{j-1})$ that the future depends on the entire
string $x^i y^{j-1}$ and does not depend on any symbol further in the
past.  Our novel interpretation leads to the following {\it
nonuniform\/} generation algorithm.

{\sf
\begin{tabbing}
aaa \= aaa \= aaa \= aaa \= \kill
{\sc nonuniform-generate}($T$,$\phi$) \\
1. \> Initialize $t := 0$; $x_1^0: = \epsilon$; \\
2. \> Until $t = T$ \\
3. \> \> Pick context length $i$ in [$0,\min(t,n-1)$] \\
   \> \> \> $\pbar_c(i|x^t) = 
		\lambda(x_{t-i+1}^t) 
		\prod_{l=\min(t,n-1)}^{i+1} (1 - \lambda(x_{t-l+1}^t))$ \\
4. \> \> $c := x_{t-i+1}^t$; $\jmax := \max(n-i, T-t)$; \\
5. \> \> Pick prediction $y_1^j$ of length $j$ in [$1,\jmax$] \\
   \> \> \> $\pbar_v(y_1^j|c) = 
	(1 - \lambda(c y_1^j)) \delta(y_j|c y_1^{j-1}, i+j)
	\prod_{l=1}^{i-1} \lambda(c y_1^l)
				\delta(y_l|c y_1^{l-1}, l+i)$ \\
   \> \> \> where $\lambda(c y_1^{\jmax}) \doteq 0$. \\
6. \> \> Extend history $x_1^t$ by prediction $y_1^j$\\
   \> \> \> $x_1^{t+j} := x_1^t y_1^j$; $t := t + j$; \\
7. \> return($x^T$); \\
\end{tabbing}
}

The nonuniform model behaves both like a state model and like a
context model.  The transition from a longer context to a shorter
context (line 3) continues for the duration of the resulting
prediction (line 5).  If a unit length prediction is made, then the
nonuniform model behaves exactly like the context model.  However, if
a longer prediction is made, then the nonuniform model behaves more
like the state model.

  \subsection{Two Situations}

Let us examine the behavior of our three model classes in two
situations.  The first situation is a point of local independence,
where the current prediction does not depend on the history but later
predictions do.  In such a situation, the context model will
outperform the state model.  The second situation is a point of global
independence, where no subsequent prediction depends on the current
history.  In such a situation, the state model will outperform the
context model.  The nonuniform model will perform reasonably well in
both situations. 

\begin{figure}
\figurerule
{\small
\begin{equation}\label{context-trigram-eqn}
p_c(x^3|3,\phi) = \delta(x_1) 
	\left[ \begin{array}{l}
	  \lambda(x_1) \delta(x_2|x_1) \\
	 +(1-\lambda(x_1)) \delta(x_2)
	\end{array} \right]
	\cdot
	\left[ \begin{array}{l}
	  \lambda(x^2) \delta(x_3|x^2) \\[0.25cm]
	 +(1-\lambda(x^2)) 
		\left[ \begin{array}{l}
		  \lambda(x_2) \delta(x_3|x_2) \\
		 +(1-\lambda(x_2)) \delta(x_3)
		\end{array} \right.
	\end{array} \right.
\end{equation}

\begin{equation}\label{state-trigram-eqn}
p_s(x^3|3,\phi) = \delta(x_1)
	\left[ \begin{array}{l}
	  \lambda(x_1) \delta(x_2|x_1)
		\left[ \begin{array}{l}
		  (1-\lambda(x^2)) 
			\left[ \begin{array}{l}
			  (1-\lambda(x_2)) \delta(x_3) \\
			 +\lambda(x_2) \delta(x_3|x_2)
		        \end{array} \right. \\[0.5cm]
		 +\lambda(x^2) \delta(x_3|x^2)
	        \end{array} \right. \\[1cm]
	 +(1-\lambda(x_1)) \delta(x_2)
		\left[ \begin{array}{l}
		  \lambda(x_2) \delta(x_3|x_2) \\
		 +(1-\lambda(x_2)) \delta(x_3)
	        \end{array} \right. \\
	\end{array} \right.
\end{equation}

\begin{equation}\label{nonuniform-trigram-eqn}
p_n(x^3|3,\phi) = \delta(x_1)
	\left[ \begin{array}{l}
	  \lambda(x_1) \delta(x_2|x_1)
		\left[ \begin{array}{l}
		  (1-\lambda(x^2))
			\left[ \begin{array}{l}
			  (1-\lambda(x^2))
				\left[ \begin{array}{l}
				  (1-\lambda(x_2)) \delta(x_3) \\
				 +\lambda(x_2) \delta(x_3|x_2)
				\end{array} \right. \\[0.25cm]
			 +\lambda(x^2) \delta(x_3|x^2)
			\end{array} \right. \\[0.5cm]
		 +\lambda(x^2) \delta(x_3|x^2)
		\end{array} \right. \\[1cm]
	 +(1-\lambda(x_1)) \delta(x_2)
		\left[ \begin{array}{l}
		  \lambda(x_2) \delta(x_3|x_2) \\[0.25cm]
		 +(1-\lambda(x_2))
			\left[ \begin{array}{l}
			  \lambda(x^2) \delta(x_3|x^2) \\
			 +(1-\lambda(x^2))
				\left[ \begin{array}{l}
				  \lambda(x_2) \delta(x_3|x_2) \\
				 +(1-\lambda(x_2)) \delta(x_3)
				\end{array} \right.
			\end{array} \right.
		\end{array} \right.
	\end{array} \right.
\end{equation}
}
\caption[Trigram Evaluation]{The total probability assigned to a
string $x^3$ by the three generative interpretations of the
interpolated trigram model.  The context model is shown in
(\ref{context-trigram-eqn}), the state model in
(\ref{state-trigram-eqn}), and the nonuniform model in
(\ref{nonuniform-trigram-eqn}).}\label{trigram-figure}
\figurerule
\end{figure}

The first situation to consider is a point of local independence,
where the immediate future $y_1$ does not depend on any suffix of the
history $x^t$, while the longer term future $y_2^n$ depends on the
entire past $x^t y_1$.  In such a situation, all $\lambda(x_i^t)$ will
be close to zero, while the $\lambda(x_1^t y)$ will be close to unity.
Consequently, the context model will accurately predict $p(\cdot|x^t)$
using the empty context $\epsilon$ and then predict $p(\cdot|x^t y)$
using the full context $x^t y$.  In contrast, the state model will
transition from the $x_1^t$ context all the way to the empty context
$\epsilon$ with high probability, which then obliges it to predict
$p(\cdot|x^t y)$ using the weak context $y$.  The behavior of the
nonuniform model depends on the value of $\lambda(y)$.  If
$\lambda(y)$ is high, then the nonuniform model will behave more like
the state model, while if $\lambda(y)$ is low, then it will behave
more like the context model.

The simplest example of such a situation is an interpolated trigram
model on a string $x^3$ of length 3, where $p(\cdot|x_1) =
p(\cdot|\epsilon)$ but $p(\cdot|x^2) \neq p(\cdot|x_2)$ and
$p(\cdot|x_2) = p(\cdot|\epsilon)$.  Then $\lambda(x_1)$ and
$\lambda(x_2)$ are close to zero, while $\lambda(x^2)$ is close to
unity.  Consequently, the state model must incorrectly treat all three
symbols as being independent (\ref{local-eqn}a), while the context
model (\ref{local-eqn}b) and the nonuniform model (\ref{local-eqn}c)
are able to correctly treat $x_2$ as independent of $x_1$, while also
treating $x_3$ as dependent on both $x_1$ and $x_2$.
\begin{equation}\label{local-eqn}
\begin{array}{ll}
 a. & p_s(x^3|\phi) \approx \delta(x_1) \delta(x_2) \delta(x_3) \\
 b. & p_c(x^3|\phi) \approx \delta(x_1) \delta(x_2) \delta(x_3|x^2) \\
 c. & p_n(x^3|\phi) \approx \delta(x_1) \delta(x_2) \delta(x_3|x^2)
\end{array}
\end{equation}
The total probability assigned to a string $x^3$ by our three
interpolated trigram models appears in figure~\ref{trigram-figure}.

The second situation to consider is a point of global independence,
where the entire future $y^n$ is completely independent of the past
$x^{n-1}$.  Such a situation will arise in practice when all suffixes
of the history $x^{n-1}$ are rare, or when the source
$p(y^n|x_i^{n-1}) = p(y^n|\epsilon)$ for all $i$.  In this situation,
we would like to ignore the entire history $x^{n-1}$ when making our
predictions.  All $\lambda(x_i^{n-1})$ and
$\lambda(x_i^{n-1}y_1^{i-1})$ will be close to zero, but never
identically zero.  Due to inadequate statistics at a point of
independence, nearly all $\delta(y_1^{i-1}|x_i^{n-1})$ will be zero,
and to simplify the example we assume that all are zero.

Once the state model transitions to the empty context in order to
predict the first symbol $y_1$, it need never again transition past
any suffix of $x^{n-1}$.  The total probability assigned to
$p(A^n|\epsilon)$ by the state model (\ref{global-eqn}a) is a product
of $n-1$ probabilities.  In contrast, the context model must
transition past some suffix of the history $x^{n-1}$ for each of the
next $n-1$ predictions, and so the total probability assigned to
$p(A^p|\epsilon)$ by the context model (\ref{global-eqn}b) is a
product of $n(n-1)/2$ probabilities.  Note that (\ref{global-eqn}b)
must be considerably less than (\ref{global-eqn}a).  Here the
nonuniform model behaves like the state model by first transitioning
to the empty context and then predicting the string $y_1^n$ of length
$n$, and so the total probability assigned to $p(A^n|\epsilon)$ by the
nonuniform model (\ref{global-eqn}c) is a product of only $2n-2$
probabilities. Therefore the total probability assigned to $y^n$ by
the nonuniform model is considerably greater than that assigned by the
context model (\ref{global-eqn}b) and only slightly less than that
assigned by the state model (\ref{global-eqn}a).
\begin{equation}\label{global-eqn}
\begin{array}{ll}
 a. & \prod_{i=1}^{n-1} (1-\lambda(x_1^q)) \\
 b. & \prod_{k=1}^{n-1} \prod_{i=k}^{n-1} (1-\lambda(x_i^{n-1}y_1^{i-1})) \\
 c. & [\prod_{i=1}^{n-1} (1-\lambda(x_1^q))] [\prod_{j=1}^{n-1} \lambda(y_1^j))] \\
\end{array}
\end{equation}

The point of these examples has been to illustrate how the nonuniform
model combines the best characteristics of the state and context
models.  Like the state model, it can effectively ignore a misleading
history.  And like the context model, it does not get tricked by
points of local independence.

  \subsection{Inequivalence}

The only difference between the three model classes is how they
interpret the $\lambda$ parameters.  This raises the question of
whether the nonuniform interpretation has substance, that is, whether
every nonuniform model might really be equivalent to some uniform
model.  Here we argue that nonuniform models are fundamentally
different from uniform models, because it is not possible to map a
nondegenerate nonuniform model into an extensionally equivalent
context model or state model (theorem \ref{inequivalence-theorem}).

We say an interpolated model is {\it degenerate\/} iff it is
equivalent to some simpler model, that is, equivalent to a model with
fewer parameters.  Formally, an interpolated Markov model $\phi =
\tuple<n,A,\delta,\lambda>$ is degenerate iff either (i) some
$\lambda$ value is either 0 or 1 or (ii) some higher order transition
probability is equivalent to a lower order transition probability,
ie., $\delta(x_{i+1}|x_1^i) = \delta(x_{i+1}|x_2^i)$ for some $x_1^i$
in $A^{<n}$.

\begin{theorem}\label{inequivalence-theorem}
For every nondegenerate $\phi = \tuple<n,A,\delta,\lambda>$ and 
$\phi^\prime = \tuple<n,A,\delta,\lambda^\prime>$, with $n > 1$, there
exist strings $x^i\in A^*$ and $y^j\in A^+$ such that the nonuniform
probability $p_n(y^j|x^i,\phi)$ is not equal to the context model
probability $p_c(y^j|x^i,\phi^\prime)$ or the state model probability
$p_s(y^j|x^i,\phi^\prime)$.
\end{theorem}
\proof 
Either $\lambda = \lambda^\prime$ or $\lambda\neq\lambda^\prime$.  

{\it Case i.\/}
If $\lambda\neq\lambda^\prime$, then 
$p_n(y^1|x^i,\phi) \neq p_c(y^1|x^i,\phi^\prime)$ 
and $p_n(y^1|x^i,\phi) \neq p_s(y^1|x^i,\phi^\prime)$ 
for some $x^i\in A^+$ because all three
interpretations of $\phi$ are trivially identical
for all one symbol predictions $y^1\in A$.

{\it Case ii.\/}
Otherwise $\lambda = \lambda^\prime$, and then 
$p_n(y^1|x^i,\phi) = p_c(y^1|x^i,\phi^\prime) = p_s(y^1|x^i,\phi^\prime)$
for all $x^i$ and $y^1$.  However, now it is
straightforward to show that 
$p_n(y^j|x^i,\phi) \neq p_c(y^j|x^i,\phi^\prime)$ 
and $p_n(y^j|x^i,\phi) \neq p_s(y^j|x^i,\phi^\prime)$ 
for some $x^i$ and $y^j\in A^j$ with $j > 1$.
We consider the simplest nondegenerate situation, which is $n = 2$, $j
= 2$, and $i = 0$.  This corresponds a bigram model predicting two
symbols using an empty context.  In this situation, the state model
and the context model both assign the same uniform probability
$p_u(y^2|\phi)$ to $y^2$. 
Then 
\begin{displaymath}
\begin{array}{l}
p_u(y^2|\phi) = \delta(y_1) 
		[\lambda(y_1) \delta(y_2|y_1)
		 + (1-\lambda(y_1)) \delta(y_2)] \\
p_n(y^2|\phi) = \delta(y_1) 
		[\lambda(y_1) \delta(y_2|y_1)
		 + (1-\lambda(y_1)) [ \lambda(y_1) \delta(y_2|y_1)
					+ (1-\lambda(y_1)) \delta(y_2)]] \\
\end{array}
\end{displaymath}
and
\begin{equation}\label{delta-equation}
p_n(y^2|\phi) - p_u(y^2|\phi) = 
	\lambda(y_1) (1-\lambda(y_1)) 
	\delta(y_1) [ \delta(y_2|y_1) - \delta(y_2) ] 
.
\end{equation}
By the definition of degeneracy, neither $\lambda(y_1)$,
$1-\lambda(y_1)$, nor $\delta(y_2|y_1) - \delta(y_2)$ can be zero.  By
the axioms of probability, some $y_1$ must have nonzero probability,
which means that $\delta(y_1)$ must be nonzero for that $y_1$, and
therefore equation (\ref{delta-equation}) must also be nonzero
for that $y_1$.
\qed

It is instructive to note that the difference (\ref{delta-equation})
between the uniform and nonuniform interpretations of a given $\phi$
is proportional to the difference between the conditional probability
$\delta(y_2|y_1)$ and the marginal probability $\delta(y_2)$.  If
$y_2$ and $y_1$ are truly independent, then with high probability
$\delta(y_2|y_1) \approx \delta(y_2)$ in our training corpus and both
interpretations assign essentially the same probability to $y^2$,
regardless of our beliefs about conditional independence.  If,
however, $y_2$ truly depends on $y_1$ then with high probability
$\delta(y_2|y_1) \neq \delta(y_2)$ in our training corpus and the
difference between the context model interpretation and the nonuniform
model interpretation depends principally on our beliefs of conditional
independence.  This difference is maximized for $\lambda(y_1) = 0.5$,
ie., when we are maximally uncertain, and vanishes when $\lambda(y_1)$
approaches 0 or 1, ie., as our certainty grows.

\section{Nonuniform Algorithms}\label{algorithms-section}

Having defined the class of nonuniform models, and compared them to
the two uniform models, let us now consider how we might effectively
use the nonuniform model class in practice.  Here we provide efficient
algorithms to evaluate the probability of a string according to a
nonuniform model (section \ref{evaluate-section}), to find the most
likely generation path for a string according to a nonuniform model
(section \ref{decode-section}), and to optimize the parameters of a
nonuniform model on a training corpus (section
\ref{estimate-section}).

 \subsection{Evaluation}\label{evaluate-section}

The nonuniform model $\phi$ assigns probability to generation paths
paired with the strings that they generate.  A string may have more
than one generation path, and so the marginal probability of a string
$x^T$ is determined by summing the joint probabilities over all
generation paths $s$.
\begin{displaymath}
	p(x^T|\phi,T) = \sum_s p(x^T, s | \phi, T)
\end{displaymath}
There are only polynomially many generation paths for a given string.

The following dynamic programming algorithm evaluates the probability
of a string $x^T$ of length $T$ in $O(n^2 T)$ time and $O(T)$ space.
The space requirements of the algorithm may be reduced to $O(n)$ at a
slight expense in clarity.  Note that $\lambda(x_{t-i+1}^{t+\jmax}) =
0$ for $\jmax = \min(T-t,n-i)$.

{\sf
\begin{tabbing}
aaa \= aaa \= aaa \= aaa \= \kill
{\sc nonuniform-evaluate}($x^T$,$\phi$) \\
1. \> For $t = 2$ to $T$ [ $\alpha_t := 0$ ]; $\alpha_1 := 1$; \\
2. \> For $t = 1$ to $T-1$ \\
3. \> \> $p_c = 1$; \\
4. \> \> for $i = \min(t,n-1)$ to $0$\\
5. \> \> \> $\pbar_c := \lambda(x_{t-i+1}^t) p_c$; $p_v := 1$; \\

6. \> \> \> for $j = 1$ to $\min(T-t,n-i)$\\
7. \> \> \> \> $\pbar_v := (1-\lambda(x_{t-i+1}^{t+j}))
				\delta(x_{t+1}^{t+j}|x_{t-i+1}^t,i+j) 
				p_v$; \\
8. \> \> \> \> $\alpha_{t+j} := \alpha_{t+j} + \alpha_t\pbar_c\pbar_v$; \\
9. \> \> \> \> $p_v := \lambda(x_{t-i+1}^{t+j}) p_v$; \\
10. \> \> $p_c := (1 - \lambda(x_{t-i+1}^t))$; \\
11. \> return($\alpha_T$); \\
\end{tabbing}
}

The $\alpha_t$ variable stores the total probability $p(x^t|\phi,t)$
for the substring $x^t$.

 \subsection{Decoding}\label{decode-section}

Decoding a string $x^T$ with respect to an nonuniform model $\phi$ is
the process of finding the single most likely generation path for that
string.  This computation is performed in $O(n^2 T)$ time and $O(T)$
space by the following dynamic programming algorithm.

{\sf
\begin{tabbing}
aaa \= aaa \= aaa \= aaa \= \kill
{\sc nonuniform-decode}($x^T$,$\phi$) \\
1. \> For $t = 2$ to $T$ [ $\alpha_t := 0$ ]; $\alpha_1 := 1$; \\
2. \> For $t = 1$ to $T-1$ \\
3. \> \> $p_c = 1$; \\
4. \> \> for $i = \min(t,n-1)$ to $0$\\
5. \> \> \> $\pbar_c := \lambda(x_{t-i+1}^t) p_c$; $p_v := 1$; \\

6. \> \> \> for $j = 1$ to $\min(T-t,n-i)$\\
7. \> \> \> \> $\pbar_v := (1-\lambda(x_{t-i+1}^{t+j}))
				\delta(x_{t+1}^{t+j}|x_{t-i+1}^t,i+j) 
				p_v$; \\
8. \> \> \> \> if ($\alpha_t\pbar_c\pbar_v > \alpha_{t+j}$)
		then [ $s_{t+j} := \tuple<i,j>$;
			$\alpha_{t+j} := \alpha_t\pbar_c\pbar_v$; ] \\
9. \> \> \> \> $p_v := \lambda(x_{t-i+1}^{t+j}) p_v$; \\
10. \> \> $p_c := (1 - \lambda(x_{t-i+1}^t))$; \\
11. \> $\shat := \phi$; $t := T$; \\
12. \> while ($t > 1$) [ $\shat := s_t \shat$; $t := t - s_{t,1}$; ] \\
13. \> return($\shat$); \\
\end{tabbing}
}

The $\alpha_t$ variable stores the probability of the most likely
generation path for $x^t$, while the $s_t$ variable stores the last
transition in the most likely generation path for $x^t$.  Each
transition in the nonuniform model is a pair $\tuple<i,j>$ indicating
that a context of length $i$ was used to make a prediction of length
$j$.  The elements $i$ and $j$ of the pair $s_t = \tuple<i,j>$ are
identified by the notation $s_{t,0}$ and $s_{t,1}$, respectively.
The most likely generation path is stored in the $\shat$ variable.

 \subsection{Estimation}\label{estimate-section}

In this section, we formulate an expectation maximization (EM)
algorithm for the nonuniform Markov model.  Our development follows
the traditional lines established for the hidden Markov model
\cite{baum-eagon:67,baum-etal:70}.  (See \cite{rabiner-juang:86} for a
tutorial.)  Recall that we must first calculate the expected number of
times that each hidden event occurred for a given training sequence.
The hidden events for the nonuniform model are the choice of context
and prediction lengths.

We begin by defining our forward and backward variables.
The forward variable $\alpha_t(i,j)$ contains the probability of
generating the first $t$ symbols of the history, picking a context of
length $i$ and then making a prediction of length $j$, according to
the model $\phi$.
\begin{equation}
\alpha_t(i,j) \doteq p(h=x_1^t, c=x_{t-i+1}^t, v=x_{t+1}^{t+j} | \phi, T)
\end{equation}
The following iterative algorithm calculates all $\alpha_t(i,j)$
values in $O(n^2 T)$ time and $O(n^2 T)$ space.

{\sf
\begin{tabbing}
aaa \= aaa \= aaa \= aaa \= \kill
{\sc forward}($x^T$,$\phi$) \\
1. \> For $j = 1$ to $n$ [ $\alpha_0(0,j) := \pbar_v(x_1^j | \epsilon)$; ]; \\
2. \> For $t = 1$ to $T$ \\
3. \> \> $\alpha_t := \sum_{j=1}^{\min(t,n)}
			\sum_{i=0}^{\min(n-j,t-j)}
			\alpha_{t-j}(i,j)$; \\
4. \> \> For $i = 0$ to $\min(t,n-1)$ \\
5. \> \> \> For $j = 1$ to $\min(T-t,n-i)$ \\
6. \> \> \> \> $\alpha_t(i,j) := \alpha_t 
				\pbar_c(i|x_1^t)
				\pbar_v(x_{t+1}^{t+j}|x_{t-i+1}^t)$; \\
\end{tabbing}
}

The backward variable $\beta_t(i,j)$ contains the probability of
generating the final $T-t$ symbols in the string $x_1^T$, given that
the history is $x_1^t$ and that we have chosen to make a prediction of
length $j$ in a context of length $i$ according to the model $\phi$.
\begin{equation}
\begin{array}{rcl}
\beta_t(i,j) & \doteq & p(x_{t+1}^T | h=x_1^t, c=x_{t-i+1}^t, v=x_{t+1}^{t+j} | \phi,T) \\
		& = & p(x_{t+j+1}^T | x_1^{t+j}, \phi) = \beta_{t+j} \\
\end{array}
\end{equation}
The following iterative algorithm calculates all $\beta_t$ values in
$O(n^2 T)$ time and $O(T)$ space.  Note that we need only maintain
a one dimensional table of $\beta$ values because $\beta_t(i,j) =
\beta_{t+j}$ for all $i,j$.

{\sf
\begin{tabbing}
aaa \= aaa \= aaa \= aaa \= \kill
{\sc backward}($x^T$,$\phi$) \\
1. \> $\beta_T := 1$; \\
2. \> For $t = T-1$ to $0$ \\
3. \> \> $\beta_t := \sum_{i=0}^{\min(t,n-1)}
			\sum_{j=1}^{\min(T-t,n-i)}
			\pbar_c(i|x_1^t)
			\pbar_v(x_{t+1}^{t+j}|c=x_{t-i+1}^t)
			\beta_{t+j}$; \\
\end{tabbing}
}

The forward and backward variables allow us to calculate the posteriori
probability of every hidden transition in our model, as represented by
the following $\gamma_t(i,j)$ variable.
\begin{equation}
\begin{array}{rcl}
\gamma_t(i,j) & \doteq & p(c=x_{t-i+1}^t, v=x_{t+1}^{t+j} | x_1^T, \phi) \\
		& = & \alpha_t(i,j)\beta_t(i,j) / p(x_1^T|\phi)
		= \alpha_t(i,j)\beta_{t+j} / p(x_1^T|\phi) \\
\end{array}
\end{equation}

We use the following useful fact to verify our implementation of the
$\gamma$ computation.
\begin{theorem}
The following constraint holds for the $\gamma$ values:
\begin{equation}
T = \sum_{t=0}^{T-1} 
	\sum_{i=0}^{\min(t,n-1)}
	\sum_{j=1}^{\min(T-t,n-i)}
		j\cdot\gamma_t(i,j)
\end{equation}
\end{theorem}
\proof 
Recall that $\gamma_t(i,j)$ represents the posteriori
probability that the nonuniform model made a prediction of length $j$
using a context of length $i$ at time $t$ in the input string $x^T$.
Each such stochastic transition consumes exactly $j$ symbols of the
input.  Consequently, summing the $\gamma_t(i,j)$ over the prediction
lengths $j$ multiplied by the prediction lengths $j$ yields the
expected number of symbols predicted at time $t$ from a context of
length $i$.  Summing this quantity over the context lengths $i$ yields
the the expected number of symbols predicted at time $t$, independent
of context length.  Finally, summing this expectation over all the
times $t$ must yield the total number of symbols in a string $x^T$.
\qed

We sum the $\gamma$ values to obtain the expected number of times that
the nonuniform model transitioned from a longer context to a shorter
one, or from a shorter prediction to a longer one.  We use two
variables to keep track of our expectations: $\lambda^+(y^l)$
accumulates the number of times that we used $y^l$ to condition our
prediction when it was possible to do so, while $\lambda^-(y^l)$
accumulates the number of times that we could have used $y^l$ to
condition our prediction but chose a proper suffix instead.  The
following algorithm accumulates all $\lambda^+(y^l)$ and
$\lambda^-(y^l)$ values in $O(n^3 T)$ time and $O(n^2 T)$ space.

{\sf
\begin{tabbing}
aaa \= aaa \= aaa \= aaa \= \kill
{\sc expectation-step}($x^T$,$\phi$,$\lambda^+$,$\lambda^-$) \\
1. \> For $t = 1$ to $T$ \\
2. \> \> For $i = 0$ to $\min(t,n-1)$ \\
3. \> \> \> For $j = 1$ to $\min(T-t,n-i)$ \\
4. \> \> \> \> $\lambda^+(x_{t-i+1}^t) += \gamma_t(i,j)$; \\
5. \> \> \> \> $\lambda^-(x_{t-i+1}^{t+j}) += \gamma_t(i,j)$; \\
6. \> \> \> \> For $l = i+1$ to $\min(t,n-1)$ 
	[ $\lambda^-(x_{t-l+1}^t) += \gamma_t(i,j)$; ]; \\
7. \> \> \> \> For $l = j+1$ to $\min(T-t,n-i)$ 
	[ $\lambda^+(x_{t-i+1}^{t+l}) += \gamma_t(i,j)$; ]; \\
\end{tabbing}
}

Having done all the work in the expectation step, the maximization
step is straightforward.

{\sf
\begin{tabbing}
aaa \= aaa \= aaa \= aaa \= \kill
{\sc maximization-step}($\phi$,$\lambda^+$,$\lambda^-$) \\
1. \> For all strings $y^l$ in $A^{<n}$ \\
2. \> \> $\bar{\lambda(y^l)} := 
	\lambda^+(y^l) / (\lambda^+(y^l) + \lambda^-(y^l))$; \\
\end{tabbing}
}

The following {\sc deleted-estimation\/}() algorithm estimates the
parameters of an interpolated model $\phi$ using a set ${\bf B}$ of
blocks of text.  For each iteration, we delete one block $B_i$ from
the set ${\bf B}$, initialize the string probabilities $\delta$ to
their empirical probabilities in the remaining blocks ${\bf B} - B_i$
(line 4), and then perform an expectation step on the deleted block
$B_i$ (line 5).  After all blocks have been deleted, we update our
model parameters (line 6).

{\sf
\begin{tabbing}
aaa \= aaa \= aaa \= aaa \= \kill
{\sc deleted-estimation}(${\bf B}$,$\phi$) \\
1. \> Until convergence\\
2. \> \> Initialize $\lambda^+,\lambda^-$ to zero;\\
3. \> \> For each block $B_i$ in ${\bf B}$ \\
4. \> \> \> Initialize $\delta$ using ${\bf B} - B_i$; \\
5. \> \> \> {\sc expectation-step\/}($B_i$,$\phi$,$\lambda^+$,$\lambda^-$); \\
6. \> \> {\sc maximization-step\/}($\phi$,$\lambda^+$,$\lambda^-$); \\
7. \> Initialize $\delta$ using ${\bf B}$; \\
\end{tabbing}
}

\section{Experimental Results}\label{empirical-section}

In this section we compare the performance of the interpolated context
model and the nonuniform model on the Wall Street Journal.  (Recall
that the interpolated context model is the classic interpolated Markov
model of Jelinek and Mercer \cite{jelinek-mercer:80}.)  We performed
two sets of experiments.  The first set of experiments was with the
6.2 million word WSJ 1989 corpus.  The goal of these initial
experiments was to better understand how initial parameter values
affect model performance.  The second set of experiments was with the
42.3 million word WSJ 1987-89 corpus.  In order to assess the possible
value of our language models to speech recognition, we used verbalized
punctuation and a vocabulary of approximately 20,000 words chosen from
both training and test sets.  Out-of-vocabulary words were mapped to a
unique OOV symbol.  In all cases, we used 90\% of the corpus for
training and 10\% for testing.  No parameter tying or parameter
selection was performed.  We report performance as test message
perplexity.

We set the $\delta$ parameters to be the empirical probabilities in
the training data and then optimized the $\lambda$ parameters on the
training data using deleted interpolation
\cite{jelinek-mercer:80,bahl-etal:91}.  We soon discovered that the
initial values for the $\lambda$ parameters had a noticeable effect on
model performance as did the block size used for deleted
interpolation.  Larger block sizes result in more conservative
estimates, which work better when the corpus is small relative to the
alphabet size and worse when the corpus is large relative to the
alphabet size.  More aggressive initial estimates for the $\lambda$
parameters give better initial performance for some model orders but
worse ultimate performance.  Regardless of how the $\lambda$
parameters were initialized or what block size was used, the
nonuniform model performed slightly better than the uniform model
under equivalent experimental conditions.

We considered three initial estimates for the $\lambda$ values:
uniform, the Jeffreys-Perks rule of succession
\cite{jeffreys:46,perks:47,krichevskii-trofimov:81}, and the natural
law of succession \cite{ristad:pu495-95}.  The uniform estimate sets
all $\lambda$ values to 0.5.  The Jeffreys-Perks rule sets
$\lambda(x^i)$ to $c(x^i)/(c(x^i)+k/2)$, for alphabet size $k$ and
string frequency $c(x^i)$.  Jeffreys-Perks is a conservative estimate,
that assigns relatively low probability to $\lambda(x^i)$.  The
natural law sets $\lambda(x^i)$ to
\begin{displaymath}
\frac{c(x^i)(c(x^i)+1) + q(x^i)(1-q(x^i))}{c(x^i)^2 + c(x^i) + 2q(x^i)}
\end{displaymath}
for string frequency $c(x^i)$ and context diversity $q(x^i)$ = $|\{y:
c(x^iy) > 0 \}|$.  The natural law is an aggressive estimate that
assigns relatively high probability to $\lambda(x^i)$.  The best
performance for higher model orders was achieved with uniform
initialization in all of our experiments.

 \subsection{WSJ 1989}

The first set of experiments was on the 1989 Wall Street Journal
corpus, which contains 6,219,350 words.  Our vocabulary consisted of
the 20,293 words that occurred at least 10 times in the entire WSJ
1989 corpus.  The goal of these initial experiments was to better
understand how initial values affect model performance.

 \subsubsection{Before Optimization}

The following table reports test message perplexities for WSJ 1989
before the $\lambda$ parameters were optimized using deleted
interpolation.  The best results for both models are obtained when the
$\lambda$ parameters are initialized uniformly.  Before optimization
the interpolated context model performs better than the nonuniform
model.

{\small
\begin{center}
\begin{tabular}{|l||r|r|r||r|r|r|} \hline
  & \multicolumn{3}{c||}{Context Model} & \multicolumn{3}{c|}{Nonuniform Model} \\ 
N & Jeffrey-Perks & Natural Law & 0.5 & Jeffrey-Perks & Natural Law & 0.5 \\ \hline\hline
2 & 284.9 & {\bf 188.2} & 215.9 & 276.8 & 197.6 & 209.6 \\
3 & 248.1 & 148.7 & {\bf 136.0} & 235.8 & 175.4 & 138.4 \\
4 & 241.6 & 155.0 & {\bf 130.0} & 229.3 & 196.3 & 138.3 \\
5 & 239.6 & 161.7 & {\bf 131.3} & 227.6 & 211.4 & 142.6 \\
6 & 238.7 & 165.7 & {\bf 132.6} & 226.9 & 219.4 & 145.2 \\
\hline
\end{tabular}
\end{center}
}

 \subsubsection{After Optimization}

The following table reports test message perplexities for WSJ 1989
after optimization via deleted interpolation.  All models were trained
using deleted interpolation with 22 blocks on the first 90\% of the
corpus and then tested on the remaining 10\% of the corpus.  The
nonuniform model slightly outperforms the context model for $n > 3$.
The best results for both models are obtained when the $\lambda$
parameters are initialized uniformly.  The nonuniform model is less
sensitive to the initial $\lambda$ estimates than the context model.

{\small
\begin{center}
\begin{tabular}{|l||r|r|r||r|r|r|} \hline
  & \multicolumn{3}{c||}{Context Model} & \multicolumn{3}{c|}{Nonuniform Model} \\ 
N & Jeffrey-Perks & Natural Law & 0.5
	& Jeffrey-Perks & Natural Law & 0.5 \\ \hline\hline
2 & 175.3 & {\bf 175.2} & {\bf 175.2} & 177.7 & 177.6 & 177.7 \\
3 & 122.1 & 121.8 & {\bf 121.2} & 121.6 & 121.6 & {\bf 121.2} \\
4 & 115.8 & 115.9 & 114.0 & 113.6 & 114.1 & {\bf 113.2} \\
5 & 114.5 & 115.4 & 112.6 & 111.9 & 113.0 & {\bf 111.4} \\
6 & 114.1 & 115.6 & 112.3 & 111.5 & 112.9 & {\bf 111.0} \\
\hline
\end{tabular}
\end{center}
}

 \subsection{WSJ 1987-89}

The second set of experiments was on the 1987-89 Wall Street Journal
corpus, which contains 42,373,513 words.  Our vocabulary consisted of
the 20,092 words that occurred at least 63 times in the entire WSJ
1987-89 corpus.  The goal of these experiments was to produce
competative results for the context model, in order to compare those
results to those achieved by the nonuniform model.  We believe that we
are the first to report WSJ 1987-89 results for full (ie., unpruned)
interpolated Markov models of higher order than trigrams.

 \subsubsection{Before Optimization}

The following table reports test message perplexities for WSJ 1987-89
before optimization via deleted interpolation.  All $\lambda$ values
were initialized uniformly.

{\small
\begin{center}
\begin{tabular}{|l||r||r|} \hline
N  & Context Model & Nonuniform Model \\ \hline\hline
2 & 198.2 & {\bf 190.1} \\ 
3 & 107.5 & {\bf 106.1} \\
4 & {\bf 97.7} & 100.4 \\
\hline
\end{tabular}
\end{center}
}

 \subsubsection{After Optimization}

The following table reports test message perplexities for WSJ 1987-89
after optimization via deleted interpolation.  All $\lambda$ values
were initialized uniformly, trained using deleted interpolation with
152 blocks on the first 90\% of the corpus, and then tested on the
remaining 10\% of the corpus.  The nonuniform model performs slightly
better than the context model for $n > 2$.

{\small
\begin{center}
\begin{tabular}{|l||r||r|} \hline
N  & Context Model & Nonuniform Model \\ \hline\hline
2 & {\bf 150.7} & 151.7 \\ 
3 & 93.4 & {\bf 93.3} \\
4 & 85.7 & {\bf 84.4} \\
\hline
\end{tabular}
\end{center}
}

\section{Conclusion}

We have proposed a nonuniform Markov model, that makes predictions of
varying lengths using contexts of varying lengths.  We argue that the
nonuniform model combines the ability of the context model to properly
model situations of local independence with the ability of the state
model to properly model situations of global independence.  We
demonstrated that the nonuniform model slightly outperforms the
interpolated context model on natural language text.  This feat is
somewhat remarkable when we consider that both models are based on the
statistics of fixed-length strings, and that both models contain
identical numbers of parameters whose values are estimated using
expectation-maximization.  The only difference between the two models
is how they combine the statistics of longer and shorter strings.

\end{document}